\documentclass{article}
\usepackage{spconf,amsmath,epsfig}

\usepackage{multirow}
\usepackage{multicol}

\let\OLDthebibliography\thebibliography
\renewcommand\thebibliography[1]{
  \OLDthebibliography{#1}
  \setlength{\parskip}{0pt}
  \setlength{\itemsep}{0pt plus 0.3ex}
}

\begin{document}\sloppy

% Example definitions.
% --------------------
\def\x{{\mathbf x}}
\def\L{{\cal L}}

\title{Structure-aware Image Inpainting with Two Parallel Streams}
\name{
 Zhilin Huang,
Chujun Qin,
Ruixin Liu,
Zhenyu Weng,
Yuesheng Zhu\footnote{Contact Author}
}
%{zerinhwang03,chujun.qin,anne\_xin,wzytumbler,zhuys\}@pku.edu.cn
\address{
Communication and Information Security Lab, Peking University\\
\{zerinhwang03,chujun.qin,anne\_xin,wzytumbler,zhuys\}@pku.edu.cn
}

\maketitle

\begin{abstract}
Recent works in image inpainting have shown that structural information plays an important role in recovering visually pleasing results. In this paper, we propose an end-to-end architecture composed of two parallel UNet-based streams: a main stream (MS) and a structure stream (SS). With the assistance of SS, MS can produce plausible results with reasonable structures and realistic details. Specifically, MS reconstructs detailed images by inferring missing structures and textures simultaneously, and SS restores only missing structures by processing the hierarchical information from the encoder of MS. By interacting with SS in the training process, MS can be implicitly encouraged to exploit structural cues. In order to help SS focus on structures and prevent textures in MS from being affected, a gated unit is proposed to depress structure-irrelevant activations in the information flow between MS and SS. Furthermore, the multi-scale structure feature maps in SS are utilized to explicitly guide the structure-reasonable image reconstruction in the decoder of MS through the fusion block. Extensive experiments on CelebA, Paris StreetView and Places2 datasets demonstrate that our proposed method outperforms state-of-the-art methods.
\end{abstract}
%
% \begin{keywords}

% \end{keywords}
%
\section{Introduction}
\label{sec:intro}

Image inpainting is a task of restoring the missing or damaged parts of images in computer vision, which has a wide range of practical applications, such as photo manipulation, filling missing contents, etc. The main challenge of image inpainting is to generate visually plausible results with reasonable structures and realistic details.

\begin{figure}
\centering
\includegraphics[width=8.5cm]{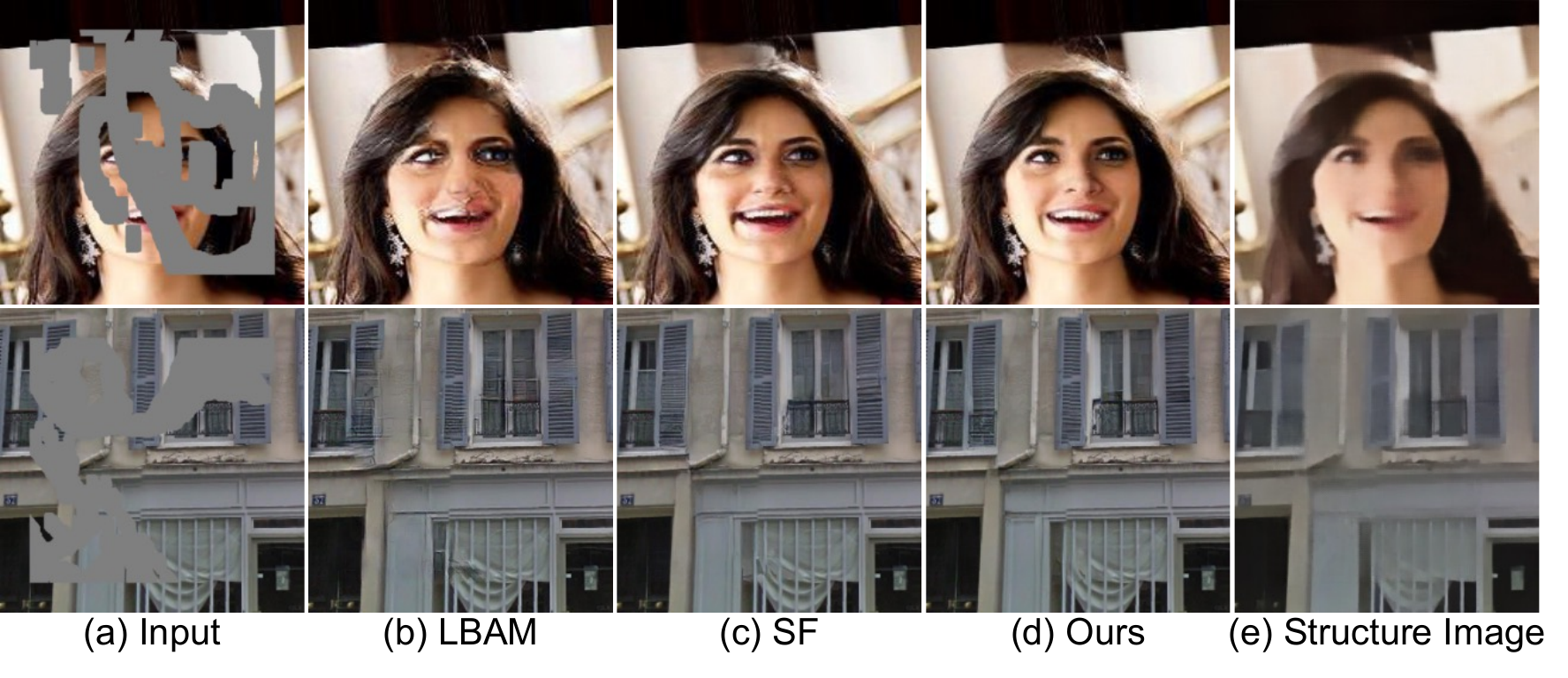}
\caption{Comparison results between LBAM, SF and ours. And (e) presents the structure images recovered by the structure stream of our model. [Best viewed with zoom-in.]}  \label{Fig_Intro}
\end{figure}

Traditionally, this task is settled with diffusion-based or patch-based approaches~\cite{DBLP:journals/tog/BarnesSFG09,DBLP:conf/iccv/EfrosL99} which extract the relevant image patches from known regions to fill the missing holes. These methods only work well for stationary regions and always fail to generate semantic information on non-stationary images since they are not able to "understand" the image. 
To make up for it, learning-based methods~\cite{DBLP:conf/cvpr/PathakKDDE16,DBLP:conf/iccv/XieLLCZLWD19,DBLP:conf/aaai/YuGJW0LZL20} are proposed to formulate inpainting as a conditional image generation problem by using a convolutional encoder-decoder network, where the encoder learns a latent feature representation of the image and then the decoder takes the representation to reason about the missing contents. However, without enough guidance, they always fail to effectively exploit structures (like edges and low-frequency information~\cite{DBLP:conf/iccv/RenYZLLL19}) and cannot predict the missing structures correctly when the holes are completely empty, leading to blurry contents or artifacts as shown in Fig.~\ref{Fig_Intro} (b) LBAM~\cite{DBLP:conf/iccv/XieLLCZLWD19}. 

To handle this problem, an end-to-end model composed of two stacked encoder-decoder networks~\cite{DBLP:conf/cvpr/Yu0YSLH18,DBLP:conf/iccv/YuLYSLH19,DBLP:conf/mmm/HuangQLLZ21,huang2021semantic} are proposed to divide the inpainting process into coarse and fine stage. They reconstruct a preliminary structure image in the coarse stage and then use it to guide the generation of detailed images in the fine stage. And ground-truth images are adopted as labels of both two stages for the training.
However, the high-frequency textures contained in ground-truth images may mislead the structure reconstruction in the coarse stage, further bringing negative effect to the subsequent detailed image reconstruction.
Recently,~\cite{DBLP:conf/icassp/LiaoH0W18,DBLP:conf/iccvw/NazeriNJQE19,DBLP:conf/iccv/RenYZLLL19} propose to adopt similar two-stage architectures like previous methods. 
They replace the ground-truth images with the structure images (like sketch maps, edge preserved smoothed images, etc.) as labels of the first stage, and optimize the two stages separately before jointly fine-tuning the two stages. 
By doing so, the first stage can focus on exploiting structural information and predicting missing structures in holes, thus obtaining a reasonable structure image to assist the subsequent detailed image reconstruction in the second stage.
These methods have made significant progress, presenting that structural information plays an important role in recovering visually pleasing results.
However, their complex training schedule requires additional training time and computation resources and sometimes they fail to recover significant structures as shown in Fig.~\ref{Fig_Intro} (c) SF~\cite{DBLP:conf/iccv/RenYZLLL19}. The reason is that
the way they utilize the structural information from the first stage is inefficient: (1) They only consider the image-level output of the first stage, while neglect the useful feature-level structural information learned in intermediate layers. (2) The pre-generated structural information which is only incorporated into the beginning of the second stage cannot provide sufficient guidance to produce structure-reasonable images in the decoder since it will be weaken as it reaches the end of the network~\cite{DBLP:conf/cvpr/HuangLMW17}.

To effectively exploit and utilize structural information in images for image inpainting, in this paper, we propose an end-to-end architecture composed of two parallel streams. We utilize a main stream (MS) to reconstruct corrupted images, and introduce a parallel structure stream (SS) to assist MS to produce visually plausible results with reasonable structures and realistic details. 
Specifically, MS aims at producing detailed images by processing structures and textures simultaneously, SS is designed to solely extract structural representations and recover missing structures by processing hierarchical information received from the encoder of MS.
To help the structural information be directed to SS, we supervise it with a structure loss. 
In addition, we propose a gated unit (GU) to control the information flow from MS to SS. By highlighting structure-relevant activations and depressing structure-irrelevant ones, GU can not only help SS focus on structures without being misled by irrelevant information, but also prevent the textures in MS from being "washed out" in the training process. 
By interacting with SS in the joint optimization of the two streams, MS can be implicitly encouraged to exploit structural cues, thus correctly modeling the relationship between structures and textures. 
Furthermore, for fully utilizing the extracted structural information in SS to guide the image reconstruction in MS, we incorporate the multi-scale structure feature maps from the decoder of SS into the corresponding layers of MS through the proposed adaptive fusion block (AFBlk). 
By adaptively adjusting the contribution of structures and textures in both spatial ("where") and channel ("what") dimensions, AFBlk is able to effectively fuse the structure information from SS into MS while preserving the textures generated in previous decoder layers of MS.

Extensive experiments on CelebA, Paris StreetView and Places2 datasets demonstrate that our method can generate plausible results with clear boundaries and realistic details compared to state-of-the-art methods, as shown in Fig.~\ref{Fig_Intro}.
The main contributions of this paper as follows:

\begin{itemize}
\item We propose an end-to-end architecture composed of two parallel streams to effectively exploit structural information in images for image inpainting. Under the help of the structure stream (SS) which is designed to solely process the structural information, the main stream (MS) is able to produce visually pleasing results.

\item The gated unit is proposed to control the information interaction between two streams in our architecture, which can not only help SS to focus on structures, but also prevent the necessary information in MS from being affected in the training process.

\item We show that our approach can achieve promising performance compared with previous state-of-the-art methods on benchmark datasets including CelebA, Paris StreetView and Places2.
\end{itemize}

% Related Work
\section{Related Work}
\subsection{Deep Image Inpainting}
Image inpainting have achieved great success by introducing deep convolutional network especially the generative adversarial network. A pioneer work, Context Encoder (CE)~\cite{DBLP:conf/cvpr/PathakKDDE16}, formulates inpainting as a conditional image generation problem by using a convolutional encoder-decoder network, where the decoder reasons about the missing contents according to the image representation extracted by the encoder. Iizuka et al.~\cite{DBLP:journals/tog/IizukaS017} use global and local discriminators to generate better results by considering both overall consistency and local details. 
Xie et al.~\cite{DBLP:conf/iccv/XieLLCZLWD19} propose a learnable attention map module to learn mask updating for irregular holes.
Yu et al.~\cite{DBLP:conf/cvpr/Yu0YSLH18,DBLP:conf/iccv/YuLYSLH19} suggest to adopt coarse-to-fine architectures for better recovering structures and textures. 
By taking the preliminary structure image generated in the coarse stage as guidance, the fine stage can produce plausible results much easier. For distinguishing valid and invalid pixels, Yu et al.~\cite{DBLP:conf/iccv/YuLYSLH19} propose a gated convolution by introducing the gated mechanism to evaluate the validness for each pixel of the input feature map in an adaptive manner. 
However, the lack of structure guidance restricts the above methods to produce plausible structures.

\begin{figure*}
\centering
\includegraphics[width=\textwidth]{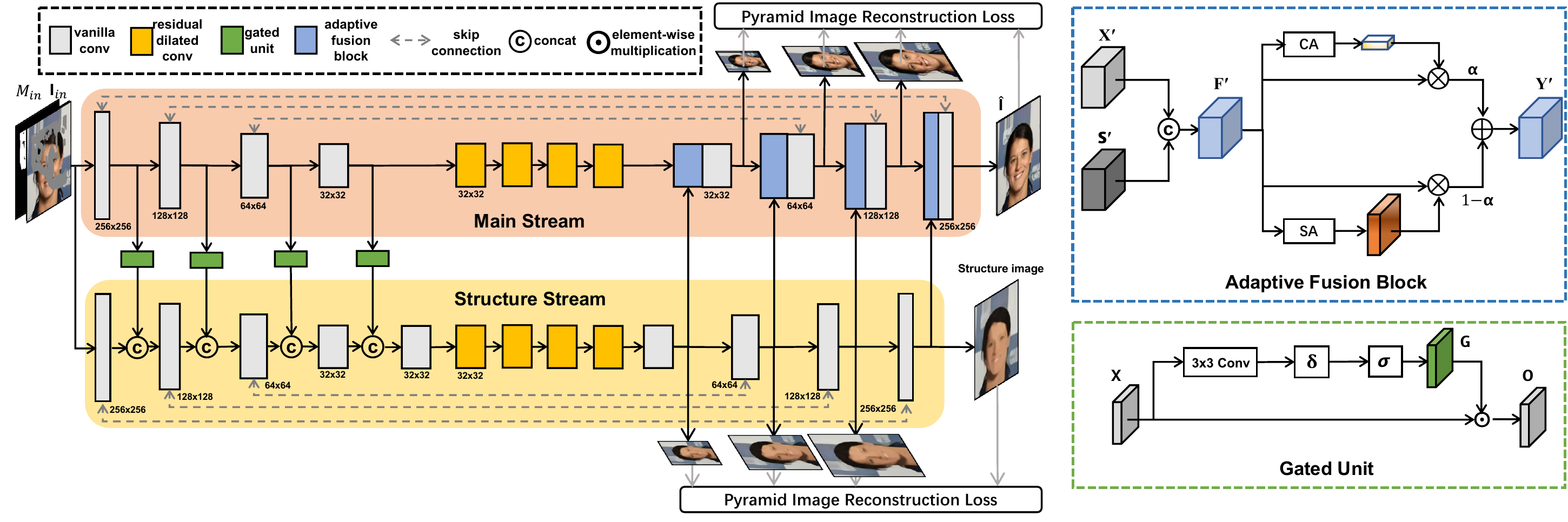}
\caption{The architecture of our proposed model.}  \label{Fig_Network}
\end{figure*}

\subsection{Structure-guided Image Inpainting}
In order to reconstruct more reasonable structures, some image inpainting methods try to take the structure knowledge as guidance. Ren et al.~\cite{DBLP:conf/iccv/RenYZLLL19} use two-stage networks and employ edge preserved smooth images as labels of the first stage. Then, the recovered structure image of the first stage is used to guide the subsequent textures generation.
Some methods~\cite{DBLP:conf/icassp/LiaoH0W18,DBLP:conf/iccvw/NazeriNJQE19,DBLP:conf/bmvc/SongYSWHK18} adopt similar architectures but utilizing the edge sketches and semantic mask as labels of the first stage, respectively.
More recently, some methods try to exploit structure knowledge in image inpainting with a multi-task framework.
Yang et al.~\cite{DBLP:conf/aaai/YangQS20} propose a shared encoder-decoder architecture to reconstruct detailed images and structures simultaneously to incorporate structure information for image inpainting. 
Liu  et al.~\cite{DBLP:conf/eccv/LiuJSHY20} propose a mutual encoder-decoder architecture which is more related to our work. They recognize features from the shallow and deep encoder layers as textures and structures, respectively, and process structures and textures separately with two branches. 
However, they neglect that directly extracting structure or texture representations from the generator which is designed to reconstruct detailed images may bring about negative effects. Specifically, since the shallow layers of the generator will receive the feedback from the texture branch during the training process, they will be implicitly enforced to focus on textures and wash out the structures, thus affecting the structure reconstruction in deep layers.
In contrast to their model, we introduce a structure stream (SS) in parallel to the main stream (MS) to process structure information separately, and apply a gated unit (GU) to the connection between MS and SS. By depressing structure-irrelevant activations in the information flow from MS to SS, GU can prevent the textures in MS from being affected in the training process.

\section{Approaches}
\subsection{Architecture}
We propose a single-stage architecture which consists of two parallel UNet-based streams: the main stream (MS) and the structure stream (SS), as shown in Fig.~\ref{Fig_Network}. MS aims at recovering missing structures and textures simultaneously, and SS is designed to solely process structural information in parallel to MS. Gated units (GU) are applied to control the information flow from MS to SS hierarchically. Then, we utilize the adaptive fusion block (AFBlk) to incorporate the multi-scale structure feature maps from decoder of SS back into the corresponding layers of MS. In the training step, the pyramid image reconstruction loss, perceptual loss are introduced to MS and SS. In addition, style loss and adversarial loss are introduced to MS. And two streams are optimized jointly. Given UNet-based architectures of $L$ layers for both MS and SS, we denote the feature maps in the encoder of MS as $X^L, X^{L-1},...,X^1$, in the decoder of MS as $X'^L, X'^{L-1},...,X'^1$, in the encoder of SS as $S^L, S^{L-1},...,S^1$, and in the decoder of SS as $S'^L, S'^{L-1},...,S'^1$, where larger number indicates smaller resolution.

\subsubsection{Main Stream} 
The main stream (MS) adopts a UNet-based architecture where both the encoder and decoder are composed of 4 vanilla convolutional layers with the kernel size of $4\times 4$. 
And we embed 4 stacked residual dilated convolutional blocks~\cite{DBLP:journals/corr/YuK15}~\cite{gottlob:nonmon} in the bottleneck to enlarge the receptive fields.
MS concatenates the corrupted image $I_{in} \in R^{3\times H \times W}$ and the mask $M_{in} \in R^{H \times W}$ in the channel dimension as input, and outputs the recovered detailed image $\hat{I} \in R^{3\times H \times W}$. Specifically, the encoder firstly reasons about missing structures and textures in holes simultaneously. Then, the AFBlk embedded in each decoder layer integrates the information from previous layer with the structure feature map from the corresponding layer of SS, for generating structure-plausible results.

\subsubsection{Structure Stream} 
The structure stream (SS) has the similar architecture with MS. The first encoder layer of SS takes the concatenation of the corrupted detailed image and the mask as input, 
and the other encoder layers of SS take the concatenation of the feature map from previous encoder layer and the corresponding layer of MS as input.
For helping SS focus on the structural information and preventing the texture information in MS from being affected in the training process, GU is applied layer-wisely to control the information flow from MS to SS.
Finally, the decoder of SS attempts to reconstruct structure feature maps at multiple scales. 

\subsection{Gated Unit}
Directly introducing SS to process the structural information in parallel to MS may face two problems: (1) Since MS contains both low-frequency structures and high-frequency textures, the structure-irrelevant information will mislead the structure reconstruction in SS. 
(2) Since SS is designed to solely process the structural information under the supervision of the structural images, the structure-irrelevant information (such as textures) in MS, which is shared with SS, will be weaken during the joint optimization of the two streams.
To address these problems, we embed gated units (GU) between MS and SS to control the information flow by highlighting the structure-relevant activations and depressing the irrelevant ones. In this way, GU can not only help SS focus on structures, but also prevent textures in MS from being affected in the training process.

We use a $3\times3$ convolution following the activation $\delta(\cdot)$ and the sigmoid function $\sigma(\cdot)$ for generating the gate map $G^l$, where the gating values in $G^l$ are between zero and one. In practice, we adopt LeakyReLU as our activation function. Then, $G^l$ is applied to the $X^l$ by element-wise multiplication to obtain the gated feature map $O^{l}$:
\begin{eqnarray}
G^l = \sigma(\delta(Conv_{3\times3}(X^l)))
\end{eqnarray}

In practice, we adopt LeakyReLU as our activation function. And then, $G^l$ is applied to the $X^l$ by element-wise multiplication to obtain the gated feature map $O^{l}$:
\begin{eqnarray}
O_{x,y}^{l} = G_{x,y}^l \odot X_{x,y}^l
\end{eqnarray}
where $\odot$ denotes element-wise multiplication, (x,y) denotes the location of pixel in the feature map.

Since the receptive field of the encoder layer increases layer by layer in the forward process, the GU applied on different positions of encoder layer can also notice different structural information in the image. For shallow encoders, GU usually pays attention to edge and contour information. For deeper encoder layers, GU can control the information passed from MS to SS as low-frequency information and high-level semantic information. 

By being applied to encoder hierarchically to process feature maps with different receptive fields, GU can more effectively transfer structural information of different levels and scales from MS to SS (edge information in shallow layers and overall semantic structure information in deep layers).

\subsection{Adaptive Fusion Block}
For effectively fusing the recovered structural information from SS to MS, we introduce an adaptive fusion block (AFBlk) inspired by CBAM~\cite{DBLP:conf/eccv/WooPLK18}, as shown in Fig.~\ref{Fig_Network}. 
First of all, $F_1$ is obtained by fusing$X'^l$ and $S'^l$ through concatenation operation.
Since $X'^l$ and $S'$ have different semantic information in channels: $X'^l$ contains more texture information and $S'^l$ contains more semantic structural information, the channel attention mechanism is applied on $F_1$ to highlight 'what' are important semantics by exploiting the inter-channel relationship. Meanwhile, for preserving the structural information (like edge etc.) in spatial dimension, the spatial attention mechanism is applied on $F_1$ to focuses on 'where' are the important parts. Finally, the two feature maps which are processed by channel attention and spatial attention mechanism are fused in an dynamic manner. The process is formulated as follows:
\begin{eqnarray}
F'^l = \delta(Conv_1([X'^l;S'^l]))
\end{eqnarray}
\begin{eqnarray}
F'^l_{ch} = CA(F'^l_1) \cdot F'^l,\hspace{3ex} F'^l_{sp} = SA(F'^l_1) \cdot F'^l
\end{eqnarray}
\begin{eqnarray}
Y'^l = F'^l_{ch} \cdot \alpha + F'^l_{sp} \cdot (1-\alpha)
\end{eqnarray}
where $Conv_1$ is a $1\times1$ convolution, $\delta$ is the activation function, $\alpha$ is a trainable scale parameter, $CA(\cdot)$ and $SA(\cdot)$ are channel attention mechanism~\cite{DBLP:journals/pami/HuSASW20} and spatial attention mechanism~\cite{DBLP:conf/eccv/WooPLK18} respectively:
\begin{eqnarray}
CA(F) = \gamma (MLP(AvgPool(F)))
\end{eqnarray}
\begin{eqnarray}
SA(F) = \gamma (Conv_5([AvgPool(F); MaxPool(F)]))
\end{eqnarray}
where $\gamma$ is the sigmoid function, $F$ is the input feature map and $Conv_5$ is a $5 \times 5$ convolution.

\subsection{Loss Functions}
To guide the optimization of the two streams in our model, we introduce pyramid image reconstruction loss, perceptual loss, style loss and adversarial loss in our model.

\subsubsection{Pyramid Image Reconstruction Loss}
We introduce the pyramid image reconstruction loss $\mathcal{L}_{py}$ proposed in~\cite{DBLP:conf/cvpr/ZengFCG19} to supervise the recovery of detailed and structure images at each scale in the decoder of MS and SS, respectively, which can help MS to progressively refine the synthesized image and enforce the SS to focus on processing structure information.
Specifically, the pyramid image reconstruction loss consists of the $\mathcal{L}_1$ distance between the RGB image synthesized from each decoder layer and the ground-truth image of the corresponding scale:
\begin{eqnarray}
\mathcal{L}_{py}=\sum_{l=1}^{L}\left\|I_{gt}^l-h(X'^l)\right\|_1 + \sum_{l=1}^{L}\left\|SI_{gt}^l-h(S'^l)\right\|_1
\end{eqnarray}
where $h$ denotes a $1\times 1$ convolution that transforms the feature map into an RGB image, $I^l_{gt}$ and $SI^l_{gt}$ indicate the detailed and structure ground-truth images scaled to the size of input feature maps.

\subsubsection{Perceptual Loss} For helping the model to capture structural information, we introduce the perceptual loss $\mathcal{L}_{per}$ following~\cite{DBLP:conf/eccv/LiuJSHY20} to the reconstructed detailed image in MS and the structure image in SS. The perceptual loss is defined on the ImageNet-pretrained VGG16:
\begin{eqnarray}
\mathcal{L}_{per}=\mathrm{E}[\sum_{i}\frac{1}{N_i}\left\| \phi_i(I_{pred})-\phi_i(I_{gt})) \right\|_1]
\end{eqnarray}
where $\phi_i$ is the feature map of $i-th$ layer of VGG-16. In practice, select the feature map from layers ReLU1\_1, ReLU2\_1, ReLU3\_1, ReLU4\_1, ReLU5\_1 following~\cite{DBLP:conf/eccv/LiuJSHY20}.

\subsubsection{Style Loss} For helping the model to preserve the style coherency, we introduce the style loss $\mathcal{L}_{sty}$ following~\cite{DBLP:conf/eccv/LiuJSHY20}. 
We compute the style loss with given feature map $C_i\times H_i\times W_i$ as follows:
\begin{eqnarray}
\mathcal{L}_{sty}=\mathrm{E_i}[\sum_{i}\left\| G^{\phi}_i(I_{pred})-G^{\phi}_i(I_{gt})) \right\|_1]
\end{eqnarray}
where $G^{\phi}_i$ is a $C_i \times C_i$ Gram matrix calculated from the given feature maps.
We only adopt the style loss to the feature maps in MS since we found that when we adopt the style loss to SS, the model performance degrades.

\subsubsection{Adversarial Loss} The adversarial training strategy is adopted in our work. We follow~\cite{DBLP:journals/tog/IizukaS017} to use SN-PatchGAN, and adopt the Relativistic Average LS adversarial loss~\cite{DBLP:conf/iclr/Jolicoeur-Martineau19} to MS for generating more realistic details. The adversarial loss of our generator is denoted as:
\begin{eqnarray}
\mathcal{L}_{adv} = -\mathrm{E}_{I_{gt}} \left[ D(I_{gt},  \hat{I} )^2\right]-\mathrm{E}_{\hat{I}} \left[(1- {D(I_{gt}, \hat{I})}^2)\right]
\end{eqnarray}

\section{Experiments}
\subsection{Experimental settings}
We evaluate our method on three datasets: CelebA~\cite{DBLP:conf/iccv/LiuLWT15}, Paris StreetView~\cite{DBLP:journals/cacm/DoerschSGSE15} and Places2~\cite{DBLP:journals/pami/ZhouLKO018}. For these datasets, we use the original train, validation and test splits. We train our model on an Nvidia RTX 2080Ti GPU and use Adam algorithm with a learning rate of $1\times 10^{-4}$ for optimization. All masks and images for training and testing are with the size of $256\times256$.
Besides, we adopt the same data augmentation such as flipping during training process as~\cite{DBLP:conf/eccv/LiuJSHY20}.
For fair comparison, we compare our method with 4 existing state-of-the-art methods: LBAM~\cite{DBLP:conf/iccv/XieLLCZLWD19}, EC~\cite{DBLP:conf/iccvw/NazeriNJQE19}, SF~\cite{DBLP:conf/iccv/RenYZLLL19}, RT~\cite{DBLP:conf/eccv/LiuJSHY20}. 
These methods all adopt parallel or series two-stage architecture.
To demonstrate the effectiveness of our model, we make comparisons between these methods and ours in filling irregular holes, which is obtained from~\cite{DBLP:conf/eccv/LiuRSWTC18}, with different hole-to-image area ratios.  

\begin{table}[!ht]
\begin{center}
\scalebox{0.8}{
\begin{tabular}{c|c|c|c|c|c|c}
\hline
\multicolumn{6}{c}{CelebA}\\
\hline
& Mask& LBAM& EC& SF& RT& Ours\\
\hline
\multirow{4}{*}{$\mathcal{L}_1^-$(\%)} 
& 10-20\% & 0.79& 0.86& 0.73& 0.89& \textbf{0.67} \\
& 20-30\% & 1.58& 1.61& 1.36& 1.70& \textbf{1.24} \\
& 30-40\% & 2.61& 2.55& 2.14& 2.68& \textbf{1.97} \\
& 40-50\% & 3.94& 3.74& 3.15& 3.90& \textbf{2.89}\\
\hline
\multirow{4}{*}{$FID^-$} 
& 10-20\% & 6.55& 6.73& 4.94& 7.53& \textbf{4.39} \\
& 20-30\% & 13.73& 12.27& 9.33& 13.25& \textbf{8.31} \\
& 30-40\% & 21.14& 17.30& 13.05& 18.47& \textbf{11.93} \\
& 40-50\% & 31.77& 22.02& 16.78& 22.55& \textbf{15.46}\\
\hline
\multirow{4}{*}{$SSIM^+$} 
& 10-20\% & 0.975& 0.975& 0.980& 0.973& \textbf{0.983} \\
& 20-30\% & 0.945& 0.947& 0.961& 0.944& \textbf{0.965} \\
& 30-40\% & 0.899& 0.906& 0.931& 0.903& \textbf{0.939} \\
& 40-50\% & 0.834& 0.848& 0.890& 0.850& \textbf{0.902} \\
\hline
\multirow{4}{*}{$PSNR^+$} 
& 10-20\% & 32.29& 32.05& 33.30& 31.59& \textbf{33.99} \\
& 20-30\% & 28.33& 28.48& 29.83& 27.93& \textbf{30.49} \\
& 30-40\% & 25.42& 25.76& 27.19& 25.40& \textbf{27.74} \\
& 40-50\% & 22.95& 23.46& 24.85& 23.26& \textbf{25.39} \\
\hline
\hline
\multicolumn{6}{c}{Paris StreetView}\\
\hline
& Mask& LBAM& EC& SF& RT& Ours\\
\hline
\multirow{4}{*}{$\mathcal{L}_1^-$(\%)} 
& 10-20\% & 1.07& 1.14& 1.08& 1.19& \textbf{0.93} \\
& 20-30\% & 1.96& 1.99& 1.89& 2.08& \textbf{1.70} \\
& 30-40\% & 2.94& 2.91& 2.78& 3.15& \textbf{2.62} \\
& 40-50\% & 4.31& 4.05& 3.95& 4.62& \textbf{3.81}\\
\hline
\multirow{4}{*}{$FID^-$} 
& 10-20\% & 18.55& 20.44& 17.47& 22.33& \textbf{13.67} \\
& 20-30\% & 35.43& 35.24& 33.90& 38.24& \textbf{25.83} \\
& 30-40\% & 52.35& 49.63& 46.85& 55.52& \textbf{38.34} \\
& 40-50\% & 73.92& 62.29& 65.64& 72.65& \textbf{51.78}\\
\hline
\multirow{4}{*}{$SSIM^+$} 
& 10-20\% & 0.953& 0.948& 0.953& 0.942& \textbf{0.960} \\
& 20-30\% & 0.909& 0.905& 0.913& 0.897& \textbf{0.922} \\
& 30-40\% & 0.848& 0.848& 0.855& 0.828& \textbf{0.866} \\
& 40-50\% & 0.769& 0.778& 0.782& 0.737& \textbf{0.797} \\
\hline
\multirow{4}{*}{$PSNR^+$} 
& 10-20\% & 30.97& 30.87& 31.44& 30.09& \textbf{32.02} \\
& 20-30\% & 27.68& 27.93& 28.51& 27.30& \textbf{28.77} \\
& 30-40\% & 25.57& 25.82& 26.30& 25.04& \textbf{26.33} \\
& 40-50\% & 23.51& 24.15& \textbf{24.42}& 22.97& 24.36 \\
\hline
\hline
\multicolumn{6}{c}{Places2}\\
\hline
& Mask& LBAM& EC& SF& RT& Ours\\
\hline
\multirow{4}{*}{$\mathcal{L}_1^-$(\%)}
& 10-20\% & 1.28& 1.15& 1.07& 1.20& \textbf{0.96} \\
& 20-30\% & 2.26& 1.99& 1.77& 2.18& \textbf{1.65} \\
& 30-40\% & 3.23& 2.87& 2.63& 3.30& \textbf{2.51} \\
& 40-50\% & 4.38& 3.95& 3.69& 4.66& \textbf{3.58} \\
\hline
\multirow{4}{*}{$FID^-$} 
& 10-20\% & 18.93& 15.94& 14.87& 20.71& \textbf{12.84} \\
& 20-30\% & 30.21& 25.48& 20.63& 33.62& \textbf{19.45} \\
& 30-40\% & 41.98& 34.23& 28.37& 44.86& \textbf{26.85} \\
& 40-50\% & 52.96& 43.24& 36.69& 55.51& \textbf{35.52} \\
\hline
\multirow{4}{*}{$SSIM^+$} 
& 10-20\% & 0.928& 0.937& 0.941& 0.931& \textbf{0.950} \\
& 20-30\% & 0.866& 0.883& 0.903& 0.867& \textbf{0.909} \\
& 30-40\% & 0.798& 0.824& 0.849& 0.794& \textbf{0.856} \\
& 40-50\% & 0.713& 0.748& 0.779& 0.701& \textbf{0.784} \\
\hline
\multirow{4}{*}{$PSNR^+$} 
& 10-20\% & 28.85& 29.68& 30.02& 28.97& \textbf{30.82} \\
& 20-30\% & 25.84& 26.78& 27.53& 25.85& \textbf{28.01} \\
& 30-40\% & 23.81& 24.69& 25.27& 23.60& \textbf{25.55} \\
& 40-50\% & 22.20& 22.99& 23.38& 20.80& \textbf{23.52} \\
\hline
\end{tabular}}
\end{center}
\caption{Quantitative comparison on CelebA, Paris StreetView and Places2 datasets in filling irregular holes with different hole-to-image area ratios. $^-$ Lower is better. $^+$ Higher is better.}
\label{Table_irr}
\end{table}

\begin{figure*}[!ht]
\centering
\includegraphics[width=\textwidth]{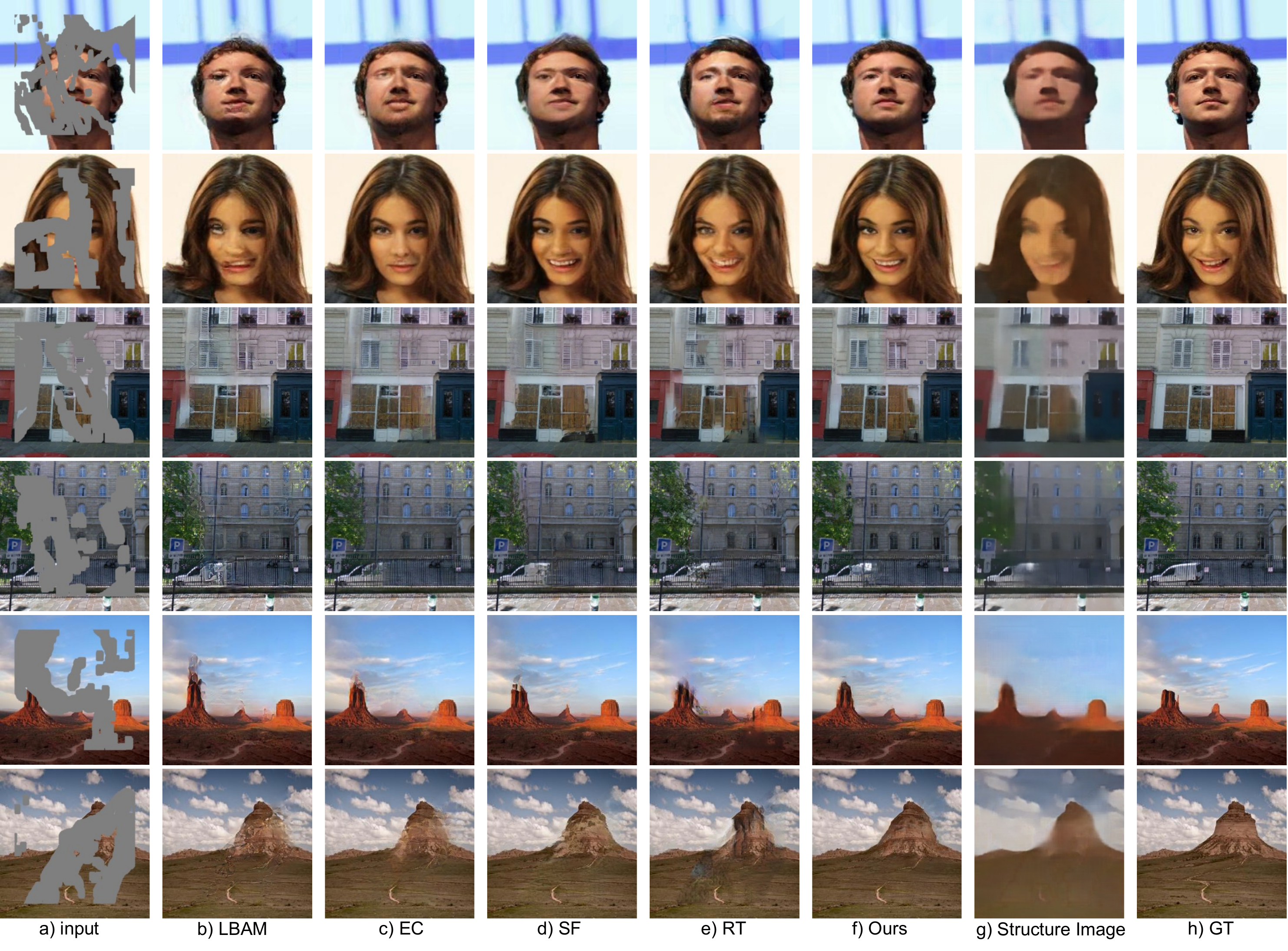}
\caption{Qualitative comparison results on CelebA, Paris StreetView and Places2 datasets with irregular holes. [Best viewed with zoom-in.]} \label{Fig_QualiComp1}
\end{figure*}

\subsection{Quantitative Comparisons}
We conduct quantitative comparisons on CelebA, Paris StreetView and Places2 datasets and we follow~\cite{DBLP:conf/aaai/YuGJW0LZL20} to choose mean $\mathcal{L}_1$ error, Peak Signal to Noise Ratio (PSNR) and Structural Similarity Index (SSIM)~\cite{DBLP:journals/tip/WangBSS04} as evaluation metrics to qualify the performance of these methods. We also introduce Fréchet Inception Distance (FID)~\cite{DBLP:conf/nips/HeuselRUNH17} to evaluate the perceptual quality of the results.
As shown in Table\ref{Table_irr}, our method outperforms all the other methods in filling irregular holes under different hole-to-image area ratios.

\subsection{Qualitative Comparisons}
We conduct qualitative comparisons on the chosen three datasets.
Fig.~\ref{Fig_QualiComp1} presents the comparison results in filling irregular holes. 
The results of LBAM and RT present that they tend to synthesize distorted structures or blurry textures without the explicit guidance of the structure information. Although results produced by EC, SF have more reasonable structures, there still exists artifacts and blurry boundaries. Compared with these methods, our model is able to synthesize visual pleasing results with clear boundaries and realistic details. Structure Image in Fig.~\ref{Fig_QualiComp1} denotes structure images generated by transforming the structure feature map $S'^1$ into a RGB image, shows that our SS is able to recover reasonable structures, thus providing reliable guidance to reconstruct detailed images in MS.

\begin{figure}[!ht]
\centering
\includegraphics[width=8cm]{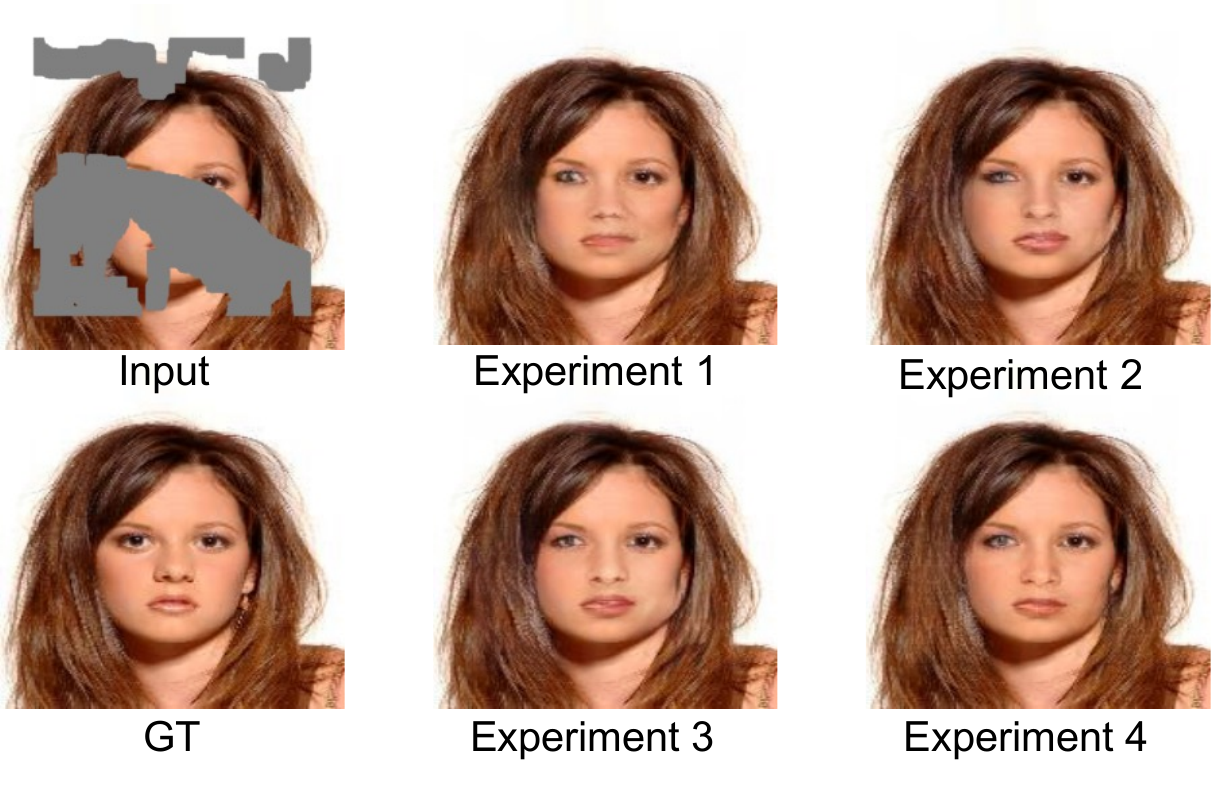}
\caption{The effect of the main components in our model. [Best viewed with zoom-in.]}  \label{Fig_maincomponent}
\end{figure}

\begin{table}[!ht]
\centering
\scalebox{0.90}{
\begin{tabular}{|l|cccc|}
\hline
\textbf{Experiment} & \textbf{$\mathcal{L}_1^-$(\%)}& \textbf{FID}$^-$& \textbf{SSIM}$^+$& \textbf{PSNR}$^+$\\
\hline
\textbf{1}: MS only & 2.21& 14.45& 0.927& 26.55\\
\hline
\textbf{2}: w/o GU & 2.01& 12.43& 0.937& 27.26\\
\hline
\textbf{3}: w/o AFBlk & 1.98& 11.99& 0.939& 27.34\\
\hline
\textbf{4}: our model& \textbf{1.92}& \textbf{11.61}& \textbf{0.941}& \textbf{27.55}\\
\hline
\end{tabular}
}
\caption{Contributions of the main components in our architecture. $^-$Lower is better. $^+$Higher is better.}%Experiments are conducted on CelebA dataset.
\label{Tab_component}
\end{table}

\subsection{Ablation Study}\label{sec_abla}
\subsubsection{The Effect of Main Components in Our Model}
We conduct five experiments on CelebA dataset to verify the effectiveness of proposed method as well as how each primary component (the structure stream, the gated unit and the adaptive fusion block) in our model contributes to the final performance.

\begin{itemize}
    \item \textbf{Experiment 1}: We removed all the other components in our model except MS.
    \item \textbf{Experiment 2}: We removed only the gated unit (GU) and kept the other components in our architecture.
    \item  \textbf{Experiment 3}: We replaced the adaptive fusion block (AFBlk) with concatenation operation and left the other components unchanged in our architecture.
    \item \textbf{Experiment 4}: Our model.
\end{itemize}

As shown in Fig.~\ref{Fig_maincomponent}, we can see that it is difficult for our model to produce satisfactory results by relying on MS alone. Comparing Experiment 2 and experiment 4, we can find that the proposed gated unit (GU) can allow SS to focus on processing structural information, enhancing the visual quality of inpainting results.
The results of Experiment 3 and 4 present that our model can generate more correct structures and realistic details with the assistance of the adaptive fusion block (AFBlk).
Similar performance on quantitative evaluation in 
Table~\ref{Tab_component} can also prove the effectiveness of each main component in our model.

\begin{figure*}[!]
\centering
\includegraphics[width=13cm]{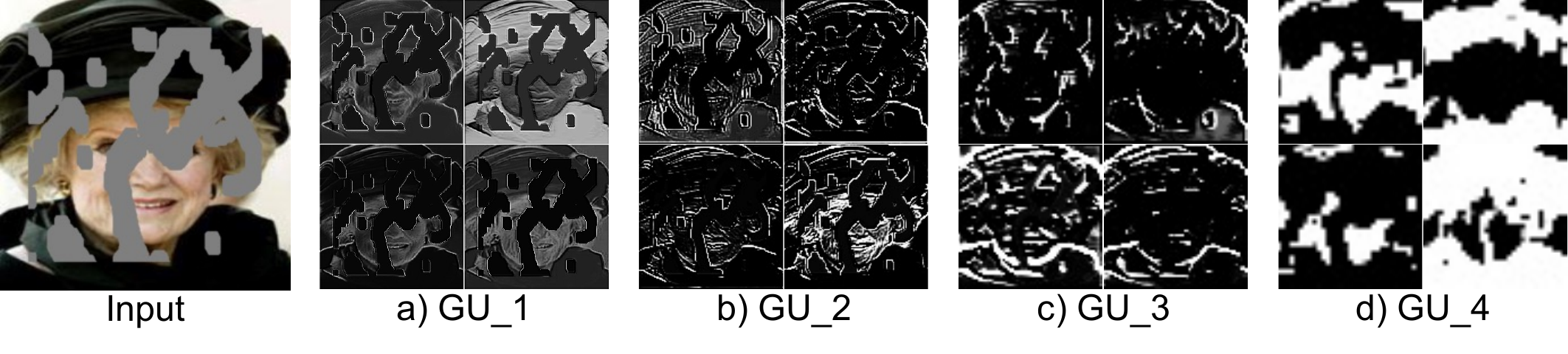}
\caption{The visualization of the GU at each level (larger number indicates smaller resolution). [Best viewed with zoom-in.]} \label{Fig_ablGU}
\end{figure*}

\begin{figure*}[!]
\centering
\includegraphics[width=13cm]{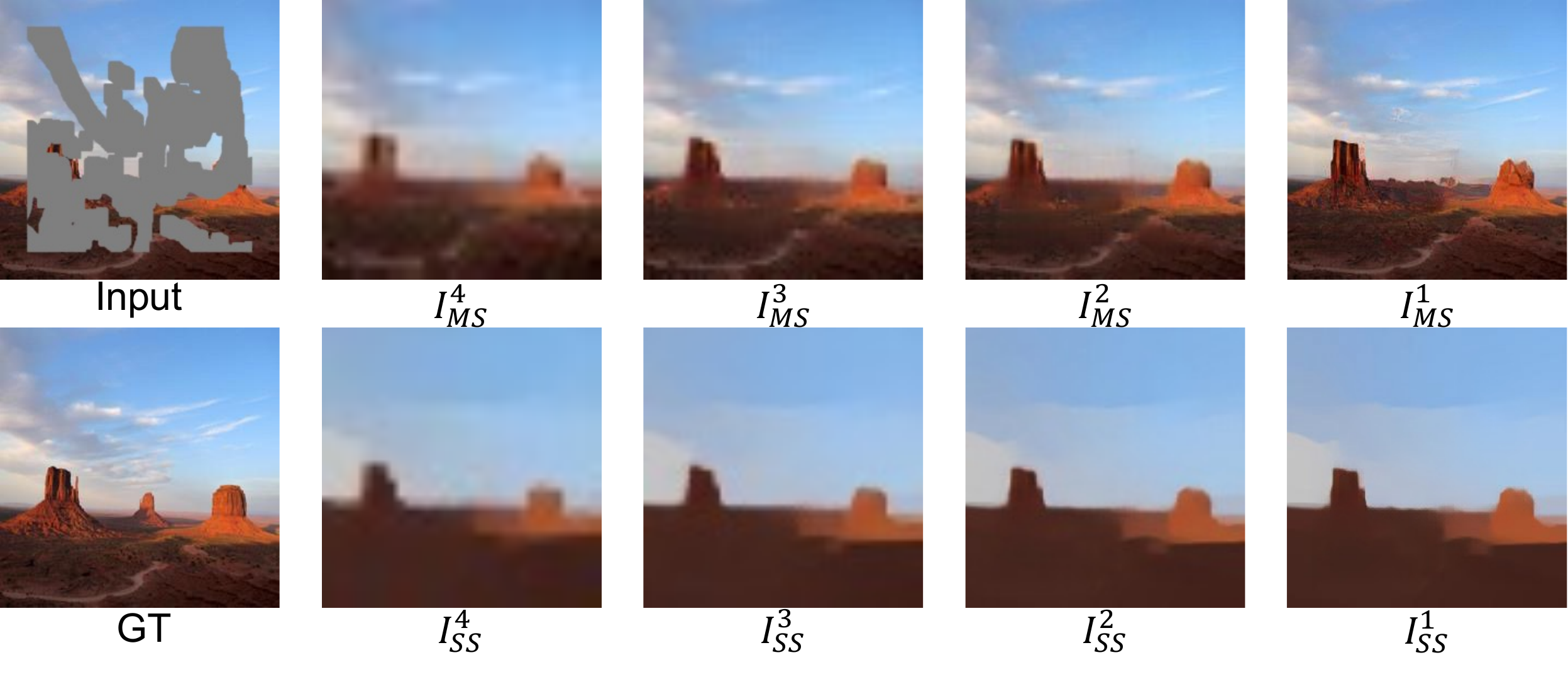}
\caption{The visualization of the detailed and structure image synthesized by feature map at each scale (larger number indicates smaller resolution). [Best viewed with zoom-in.]} \label{Fig_ablPyramid}
\end{figure*}

\subsubsection{The Visualization of GU}
We directly visualize the gate values of Gated Unit (GU) applied hierarchically to the encoder to present "what" information the GU focuses on, as shown in Fig.~\ref{Fig_ablGU}. It shows that the GU applied to shallow encoder layers can focus on the edges information and low-frequency information (like color, etc.). And the GU applied to deep encoder layers allows semantic structure information to pass from MS to SS, since high-level information extracted in corresponding layers of MS is beneficial to recover structures.

\subsubsection{The Visualization of Pyramid Detailed Images and Structure Images}
In order to demonstrate that MS in our model processes texture and structure information simultaneously, and SS only processes structure information, we present the RGB image synthesized by the feature map in each decoder layer of MS and SS in Fig.~\ref{Fig_ablPyramid}.
As shown in Fig.~\ref{Fig_ablPyramid}, we can find that the detailed images ($I_{MS}$) synthesized in decoder layer of MS contain more texture details compared with the structure images ($I_{SS}$) synthesized in the corresponding decoder layer of SS.

\subsubsection{Performance Gains with Losses}
In order to investigate the effectiveness of losses applied in MS and SS, we conduct an ablation study on Paris StreetView with randonly selected free-form masks, as shown in table~\ref{Tab_loss} and table~\ref{Tab_loss2}. The results present that applying $L_{py}$, $L_{sty}$, $L_{per}$, $L_{adv}$ to MS and applying $L_{py}$, $L_{per}$ to SS is helpful for our model to improve the final performance. In practice, we found that when we apply $L_{sty}$ on SS, SSIM and FID of the recovered detailed image generated by MS would be reduced, so we did not introduce $L_{sty}$ to SS in our model.

\begin{table}[!]
\centering
\scalebox{0.80}{
\begin{tabular}{|l|cccc|}
\hline
\textbf{Experiment} & \textbf{$\mathcal{L}_1^-$(\%)}& \textbf{FID}$^-$& \textbf{SSIM}$^+$& \textbf{PSNR}$^+$\\
\hline
$L_{py}$ & 2.68& 39.69& 0.862& 26.24\\
\hline
$L_{py}$ + $L_{per}$ & 2.59& 37.82& 0.869& 26.49\\
\hline
$L_{py}$ + $L_{adv}$ + $L_{per}$ & 2.54& 36.42& 0.872& \textbf{26.70}\\
\hline
$L_{1}$ + $L_{adv}$ + $L_{per}$ + $L_{sty}$ & 2.53& 36.51& \textbf{0.874}& 26.63\\
\hline
$L_{py}$ + $L_{adv}$ + $L_{per}$ + $L_{sty}$ & \textbf{2.51}& \textbf{36.40}& \textbf{0.874}& 26.68\\
\hline
\end{tabular}
}
\caption{Effectiveness of main losses applied to MS. $^-$Lower is better. $^+$Higher is better.}%Experiments are conducted on CelebA dataset. 
\label{Tab_loss}
\end{table}

\begin{table}[!]
\centering
\scalebox{0.90}{
\begin{tabular}{|l|cccc|}
\hline
\textbf{Experiment} & \textbf{$\mathcal{L}_1^-$(\%)}& \textbf{FID}$^-$& \textbf{SSIM}$^+$& \textbf{PSNR}$^+$\\
\hline
$L_1$ & 2.55& 36.76& 0.871& 26.59\\
\hline
$L_{py}$ & 2.54& 36.51& 0.873& 26.66\\
\hline
$L_{py}$ + $L_{per}$ & \textbf{2.51}& \textbf{36.40}& \textbf{0.874}& \textbf{26.68}\\
\hline
\end{tabular}
}
\caption{Effectiveness of main losses applied to SS. $^-$Lower is better. $^+$Higher is better.}%Experiments are conducted on CelebA dataset. 
\label{Tab_loss2}
\end{table}

\section{Conclusion}
In this paper, we propose an end-to-end architecture composed of two parallel streams to effectively exploit structural information in images for image inpainting. With the assistance of the structure stream (SS), the main stream (MS) can be encouraged to exploit structural cues, producing structure-reasonable results. A gated unit is proposed to control the information flow between MS and SS, which can help SS focus on structures and prevent textures in MS from being affected. Moreover, the multi-scale structural information in SS are utilized to explicitly guide the structure-reasonable image reconstruction in MS through the fusion block. Extensive experiments on CelebA, Paris StreetView and Places2 datasets demonstrate the superiority of our method compared with state-of-the-art methods.

\section{Acknowledgement}
This work was supported in part by the National Innovation 2030 Major S\&T Project of China under Grant 2020AAA0104203, in part by the Nature Science Foundation of China under Grant 62006007, and in part  by the Shenzhen Municipal Development and Reform Commission (Disciplinary Development Program for Data Science and Intelligent Computing).

% \bibliographystyle{IEEEbib}
% \bibliography{icme2022template}

\bibliographystyle{IEEEbib}

\end{document}